# Dynamic Knowledge Capitalization through Annotation among Economic Intelligence Actors in a Collaborative Environment

**Olusoji OKUNOYE** (*), **Fausat OLADEJO** (*), **Victor ODUMUYIWA** (*)

okunoyeo@loria.fr, oladejof@loria.fr, victor.odumuyiwa@loria.fr

(*)LORIA, Campus Scientifique, Vandoeuvre les Nancy Cedex Nancy, France

**Keywords:**
Annotation, knowledge representation, knowledge capitalisation, economic intelligence, collaboration, knowledge exploitation, group awareness

## Abstract

The shift from industrial economy to knowledge economy in today's world has revolutionalized strategic planning in organizations as well as their problem solving approaches. The point of focus today is knowledge and service production with more emphasis been laid on knowledge capital. Many organizations are investing on tools that facilitate knowledge sharing among their employees and they are as well promoting and encouraging collaboration among their staff in order to build the organization's knowledge capital with the ultimate goal of creating a lasting competitive advantage for their organizations. One of the current leading approaches used for solving organization's decision problem is the Economic Intelligence (EI) approach which involves interactions among various actors called EI actors. These actors collaborate to ensure the overall success of the decision problem solving process. In the course of the collaboration, the actors express knowledge which could be capitalized for future reuse. In this paper, we propose in the first place, an annotation model for knowledge elicitation among EI actors. Because of the need to build a knowledge capital, we also propose a dynamic knowledge capitalisation approach for managing knowledge produced by the actors. Finally, the need to manage the interactions and the interdependencies among collaborating EI actors, led to our third proposition which constitute an awareness mechanism for group work management.

# 1 Introduction

Economic Intelligence (EI) approach is a means of resolving decision problems based not only on information gathering but also on the interpretation given by the concerned actors (called EI actors) on the extracted information from available heterogeneous sources. EI concerns the set of concepts, methods, and tools which unify all the coordinated actions of research, acquisition, treatment, storage and diffusion of information, relevant to individual or clustered enterprises and organizations in the framework of strategy [9]. It is a process that involves information collection, processing and distribution with the goal of reducing uncertainty in decision making [14].

All the EI phases involve actors who collaborate to produce actionable knowledge for decision making [10] [11]. The actors of importance are mainly Decision Maker (DM), EI project coordinator and Information Watcher. They complement each others' activities in resolving a decision problem. Decision maker is the one capable of identifying and establishing the problem to be solved in terms of stake, of risk or threat on the enterprise [5]. In other words, he knows the needs of the organization, the stakes, the eventual risks and the threats the organization can be subjected to [1]. Information watcher is responsible for locating, supervising, validating, and emphasizing the strategic information needed for solving decision problem. He works hand in hand with the decision maker right from the initial stage of making a decision problem explicit. He translates this problem into information search problem so as to begin the collection of relevant information for solving the problem. EI project coordinator acts as an intermediary between all the EI actors in an organization. He is responsible for planning the work flow in EI process and coordinating the interactions among EI actors [10].

Figure 1 gives an overview of all the stages involved in EI process. For explanation purpose, EI process could be translated into five phases [10]:

- Translation phase: this phase concerns the identification of a decision problem and translating it to an information search problem. In this phase, the objectives are defined by identifying the stake in decision problem. If the decision problem is not properly identified and not properly translated then all the activities involved in the remaining four phases will be a waste [10]. Model for Decision Problem elicitation (MDP) [5] provides a way of representing stake. This model comprises of three main parameters for representing stake: the observed object, the signal and the hypothesis associated with the signal.

- Information search and retrieval phase: this is the information collection phase. After translating a decision problem into an information search problem, there is need to identify relevant information sources, validate these sources and then collect necessary information.

- Analysis phase: this is an important phase as it allows for processing collected information in order to calculate indicators necessary for decision making. These indicators are to be interpreted.

- Decision making phase: a proper interpretation of indicators will lead to better decision making.

- Protection phase: in practice, this phase is not an isolated phase because it is present throughout all the stages of EI process. For the purpose of explanation, we bring it out as a phase. Right from the decision problem identification to the final decision taken, all the expressed information and collected information should be properly protected from unwise divulgation as well as from spies and competitors. This is what we refer to as protection of information patrimony [10].

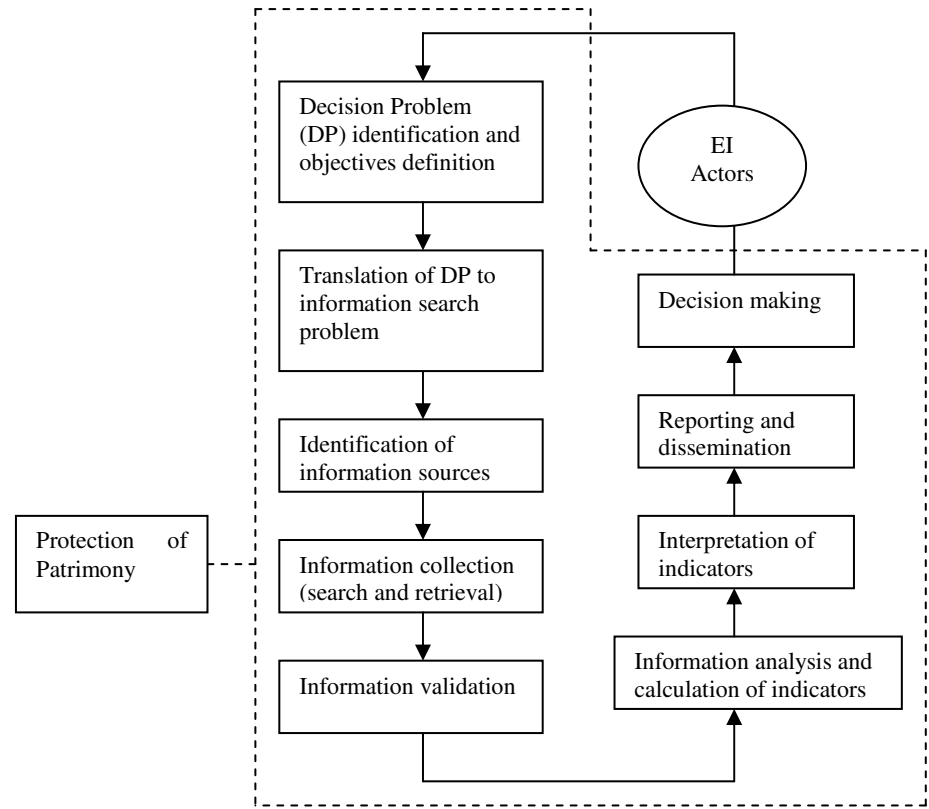

*Figure 1: Economic Intelligence process framework [10]*

EI process starts with decision maker expressing the decision problem as a form of document. This is called initial demand in [7]. Such initial demand is the materialized knowledge of the decision maker based on available information. Interactions among the actors on the initial demand generate knowledge that could give mutual and clearer understanding of the decision problem goal(s). It might involve defining or explaining some basic terms that could give a clearer picture of the perception of the decision maker, who may have formulated such decision problem, as well as identifying and understanding the goal to be achieved. Such knowledge could be captured through annotation. Shared perceptions give more insight to possible solution as actors will have clearer understanding of one another's perception. Individual may *share* their perceptions of the decision problem through *annotation* in a *collaborative* environment. The use of annotation as a means for collaborators to share one another's perception could be done synchronously or asynchronously. With synchronous annotation, users can communicate with one another in real time. It however, requires all participating members to be online. However, with asynchronous annotation, the members need not communicate in real time. A means of notifying the concerned members of pending messages may however be necessary. Knowledge produced in the course of collaboration among actors need to be capitalized for future reuse. Thus, dynamic capitalization of such knowledge is important. The questions we intend to answer in this paper are:
- How do we represent, exploit and reuse annotation in EI context?
- What is the appropriate knowledge capitalization method?
- How do we manage interactions among collaborating EI actors?

The paper is structured as follows. Section 1 discusses annotation and annotation representation for knowledge elicitation. Our proposed annotation model is discussed in section two. In section three, dynamic knowledge capitalization method is described. Section four describes the management of interactions among actors. Section five presents two systems in which the proposed model and methods were implemented. The paper is concluded in section six.

## 2 Annotation

The term annotation has been deployed and used virtually in all endeavour of life. It is an integral part of interpreting document. However, the definition and usage of the term depend on the context of use. Different users employ different annotation forms such as underlining an object of interest, writing on the margins, making a comment, use of some marks such as asterisk, question mark, etc. Annotation made may be explicit or implicit. With explicit annotation, the meaning of the annotation made is assumed to be understood at least by a community of users – users belonging to the same field of study. Unlike explicit annotation, the meaning of implicit annotation may be known only to the annotator – the person that makes the annotation.

Annotation as object is defined as an intentional and topical value-adding note linked to an extant information object [2]. Annotation is also defined as "any object (annotation) that is associated with another object (document) by some relationship" [3]. The definition of annotation by [3] does not only consider annotation as object but also as an action involving anchoring the object with the concerned document. Annotation as action is defined in [15] as an act of interpreting a document. It is a process of creating annotation as object and anchoring it to the document object (i.e. information source being annotated).

Annotation as value-added information can serve as user's interpretation of document content. It is a method of eliciting knowledge from the annotator by capturing his interpretation on the document of interest [12]. However, there is need to consider the structure such annotation takes. In existing annotation systems, users are constraint to assign values to only already defined attributes. The granularity level of attribute value a user needs may not be available. Where a user specifies an attribute that does not exist, there would be a need for structural changes to the schema of the Information System in order to accommodate such new attributes. Such changes might be tedious, time consuming and costly to effect. It is important to note that users' needs evolve and that it is impossible to pre-conceive all possible needs in the design of an information system. Our interest is to represent annotation as attribute-value pair. Annotation as attribute-value pair is a way of solving the problem of structural changes without modifying the underlying schema. Attribute-value pair in this context is user-defined. Contrary to what exists in most of the existing annotation systems, our approach allows a user to define his own attribute to depict the sense of his annotation and associate a value to the defined attribute. We assume that this approach will allow user to contextualize documents or document-objects of interest and will provide a good basis for data restructuring, data mining, robust exploitation and knowledge elicitation, among others.

In our approach to represent annotation, we identify five functional requirements that must be considered in designing an annotation system:

- Structure: the structure of an annotation is as important as the content of the annotation. As earlier explained, allowing a user to express his observation and/or contextualize document object of interest as attribute-value pair annotation could improve significantly the effectiveness of such valued-added information as opposed to adding annotation as atomic object only. Users should have the possibilities of expressing their views without restriction. They should be able to choose the word, phrase or sentence that best suit what they intend to represent.
- Communication: users may share their perceptions of the decision problem or of any information through annotation in a collaborative environment. The conclusion arrived at about the object of deliberation (decision problem) could be added to the initial information source as annotation. The use of annotation as a platform for economic actors to share one another's perception could be done synchronously or asynchronously. With synchronous annotation, users can communicate with one another in real time. It however, requires all participating actors to be online. With asynchronous annotation, the actors need not communicate in real time. However, a means of notifying the concerned actors about the pending messages is necessary.
- Anchor: annotation created can be anchored to the underlying document as inline annotation or as overlay depending on whether the user has access right to modify the document or not. Considering the wealth of information that resides in different sources of information, annotation tool that does not take into consideration the structure of documents will prove more valuable.
- Scalability: there should be the possibility of growth on annotation made. Annotations could become document for further annotations.
- Reusability: previous annotations made should be reusable. Users should be able to adapt previous annotations to similar problem without necessarily modifying it. A modified annotation becomes another annotation.
- Granularity: users should be able to add annotation to document or document object of interest at any level of granularity. No restriction should be place on what to annotate and where to place such annotation. There should be the possibility of adding annotations both at coarse-grain level and fine-grain level. For example, annotation might be added to document title, paragraph, sentence, phrase, text, word, image section etc.

Figure 2 below shows the logical view of annotation-as-process taking into consideration the complexity of annotation discussed above. The document to be annotated is fetched from document database. For a user to create annotation, he needs to log in if his profile is already stored in the user database or register as a new user. Then he can create annotation or make a follow up annotation on the document or its content. The annotation created, the anchoring information and the information on the user are stored in the annotation database. Information stored in both annotation database and user database is a form of knowledge base. Knowledge on annotation pattern of an individual actor or annotations made concerning a decision problem can be extracted. We have separated annotating the entire document from annotating parts of a document to make searching more effective. Anchoring an annotation to an entire document involves linking the annotation to the Uniform Resource Identifier of the document.

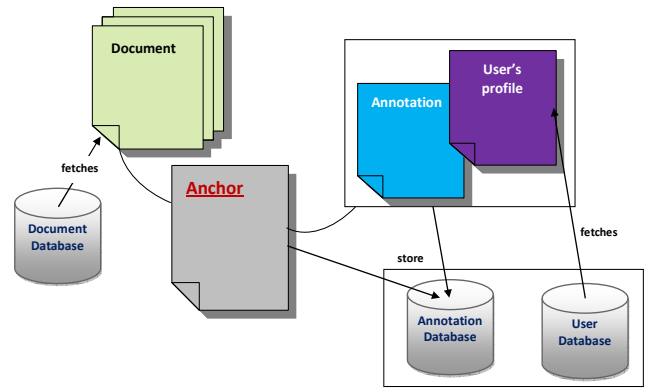

*Figure 2: Logical view of annotation-as-process*

# 3 Dynamic knowledge capitalization

Acquisition of actors' know-how and knows acquired individually or in a collaborative environment in decision making project should be such that actors would be able to capitalize reliable and up-to-date knowledge. Thus, we propose the approach of "Dynamic Capitalization" (DC) in the context of EI. We refer to the knowledge on actors, their tasks, and the results of their activities in resolution of decision problem as knowledge resources (*KR*). DC is an approach in which acquisition of *KR* covers the communication among actors in the course of resolving a decision problem. The process of validation of *KR* for reaching concession among actors is also captured with annotation process. Thus, this approach serves as the suitable methodology for KC effort in which flexible capitalization of reliable and updated knowledge is required. We present subsequently DC approach.

## 3.1 Dynamic Capitalization of Knowledge Resources

DC cycle [13] is depicted by figure 3. There are five major phases in this approach and each phase is dynamic with respect to evaluation and validation of KR by actors. Each phase is discussed in turn.

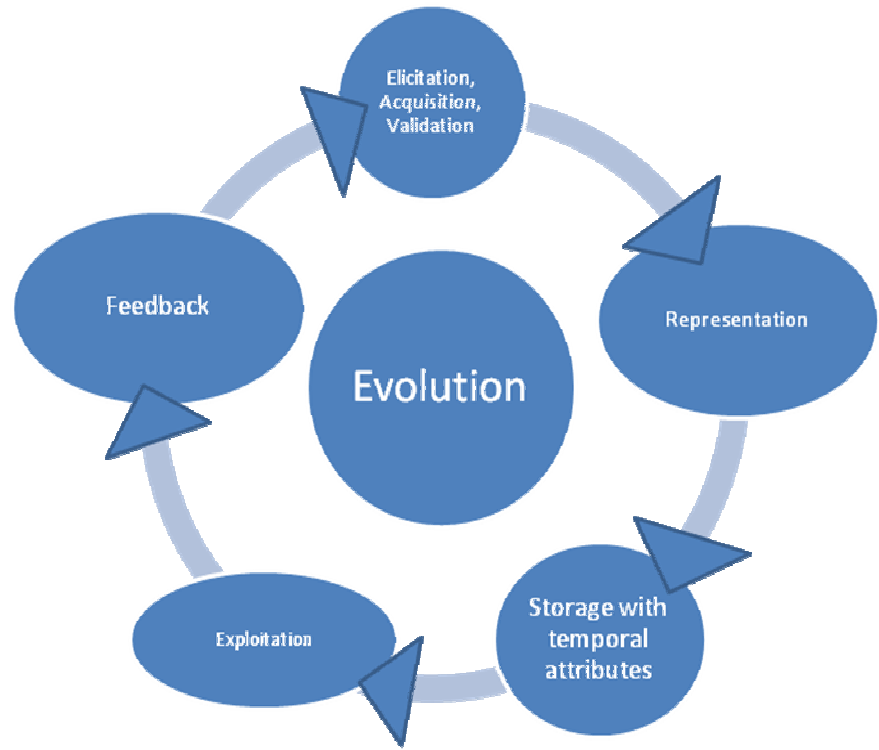

*Figure 3: DC Approach in the context of EI*

### 3.1.1 Knowledge Elicitation, Acquisition and Validation

Knowledge is elicited from actors through the process of actor-role centred acquisition. Each task with respect to the EI process is handled by respective actor, thus, their roles determines who to elicit *KR* from. The interaction among them is continuously tracked and captured. We proposed the process of declaration and annotation for this dynamic acquisition of *KR*. The declaration process allows actors to declare *KR* with respect to their roles and context at hand. For example, a decision maker expresses a decision problem which can be externalized in form of a document. The annotation process is used to capture the recurrent communication among actors. This supports the necessity of proper understanding and validation of knowledge among them. The externalized knowledge undergoes validation to ensure that there is high degree of reliability of *KR* as conceded by actors. For the implementation of this phase, the knowledge acquisition method is case-based algorithm in which roles of actors serve as the cases. Each entry of KR can be distinguished by respective timestamp such that previously captured knowledge is not replaced by new entry. Figure 4 illustrates this phase by showing the dynamic acquisition at a timestamp, denoted by $t_d$ for timestamp declaration and $t_a$ for annotation timestamp.

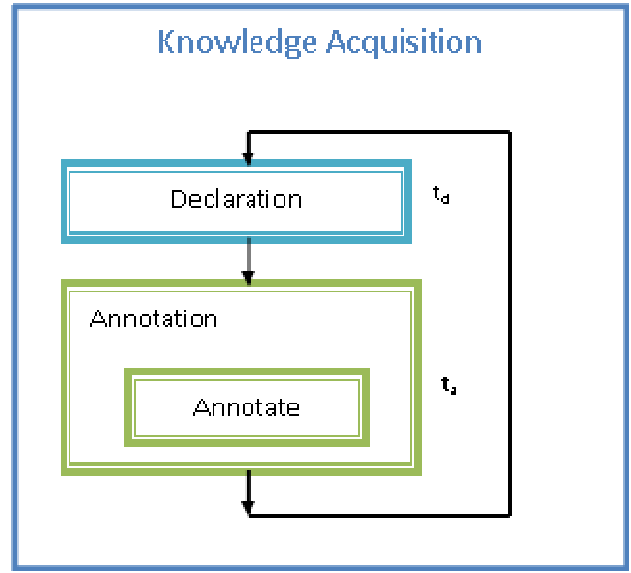

*Figure 4: Dynamic Knowledge Acquisition process*

## 3.1.2 Knowledge Representation

We represent knowledge resources with the aid of conceptual knowledge model. This models the properties of *KR* and the relationship among them. The generic conceptual model depicted in figure 5 represents general case that is applicable to problem solving process in a domain.

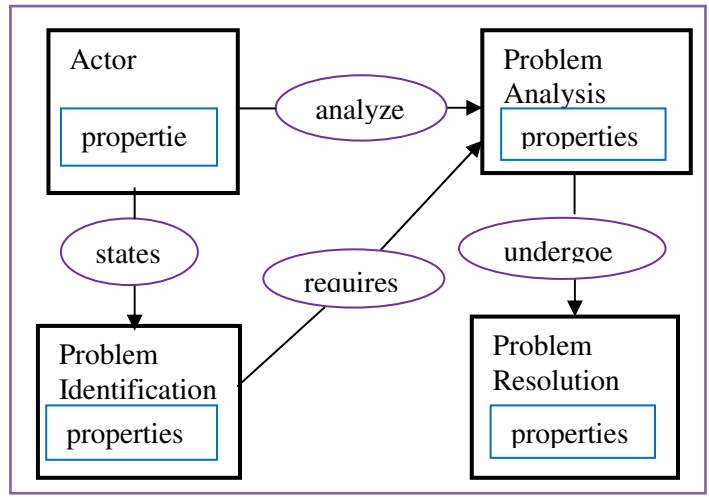

*Figure 5: Generic conceptual Knowledge Model*

### 3.1.3 Storage with Temporal Attributes

The acquired *KR* is stored with temporal attributes, that is, date/time stamp in a Knowledge Repository. The temporal attributes of *KR* facilitate dynamic and non-volatile capitalization. The storage of *KR* for instantiation of the Knowledge Repository facilitates the reuse of knowledge through exploitation process.

### 3.1.4 Exploitation of Knowledge

Acquired and stored *KR* can be exploited for reuse and sharing. EQUA²te (Explore, Query, Analyze and Annotate) model [4] is used for exploiting through query. The model guides query terms by proposing attributes of the content of Knowledge Repository to aid exploitation of relevant knowledge that meets actors' needs. Knowledge exploitation involves mining and visualization of knowledge for new cases of problems. The search algorithm adopts case-based reasoning approach for retrieval of different cases of *KR* such as, participatory actors, relevant information, result of various activities etc. from Knowledge Repository. The trend of evolution of these *KR* can be exploited as a result of the dynamic acquisition which allows storage of distinct instance of *KR* regardless of the cycle or period of its validation.

### 3.1.5 Feedback Exploitation Strategy

Actors are guided to externalize the knowledge derived from the reuse of kr in form of feedback. It includes the new problem and the assessment of the actor on the relevance of kr being exploited. Thus, it is possible to evaluate such KC system. Also, the strategy causes the Knowledge Repository to keep evolving as long as there is exploitation by actors. The method for realizing this is collaborative filtering. Thus, DC approach facilitates actors to share the lessons gained during knowledge exploitation by capturing and storing generated feedback as well as the exploited kr. This is described in figure 6.

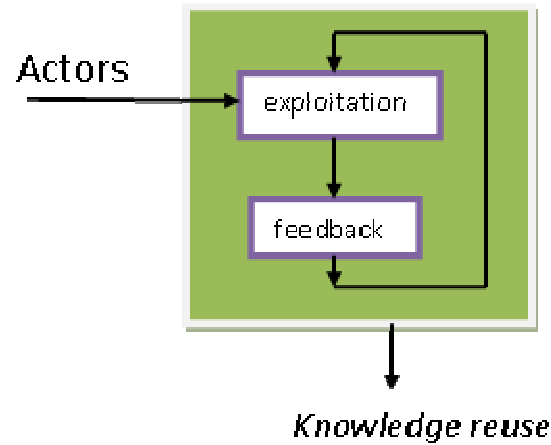

*Figure 6: Dynamic exploitation process*

Thus, the system becomes dynamic as a result of continuous update from knowledge reuse and capitalization of lesson by actors.

## 4. Managing interactions among collaborating actor

When various actors work together in a collaborative environment, there is need to coordinate their interactions. Managing interactions among collaborating EI actors is an essential element towards effective knowledge production and sharing. Our approach to managing these interactions is by enabling group awareness among the actors. When two or more persons work together on a problem; each generates a multitude of signals, either explicit or implicit. The possibility of receiving signals from each other in a group (traditional or virtual) provides an understanding of the actions and intentions of the group. In the field of Computer supported Cooperative Work (CSCW), awareness is the term used to denote the knowledge that results from the perception of signals emitted by group members [8]. Group awareness can unite stakeholders on resolving a given problem.

Awareness as defined by Dourich "*is an understanding of the activities of others, which provides a context for your own activity. This context is used to ensure that individual contributions are relevant to the group's activity as a whole and to evaluate individual actions with respect to group goals and progress.*" [6] Group awareness is achieved by capturing each actor's activities and the signals emanating from him/her in order to communicate them with his/her collaborators.

Our approach for managing the interactions among collaborating EI actors is by implementing three levels of group awareness:
- Workspace awareness: allows EI actors to have knowledge about the state of the shared workspace and its evolution. Workspace awareness can be achieved through a synchronous collaborative whiteboard or a shared editor. In an asynchronous collaboration situation, it can be achieved by facilitating workspace persistence. In the prototype we developed, we combined two approaches: shared interface and workspace persistence.
- Presence awareness: allows to know the actors that are online at any given time and their availability.
- Activity awareness: handles the communication of all activities been carried out by the different actors in the collaborative environment. It allows the actors to be aware of one another's activities in real time.

## 5. Application scenario

The propositions made in this paper have been implemented in two different systems. The first system called EIKC (Economic Intelligence Knowledge Capitalization) system allows for knowledge sharing and capitalization during decision problem identification and definition process. It also caters for the translation of the decision problem into information problem. This system allows EI actors to realize these tasks in a collaborative manner and it automatically captures the expressed knowledge for capitalization purpose.

EIKC system is simulated with a scenario of DP project stated below:

> The recent 2010/2011NECO GCE results released by National Examinations Council of Nigeria (NECO) show that 98% of candidates failed the examination[1]. This poses a major challenge to all stakeholders.

This scenario can be considered as a decision problem. Figures 7-10 show how knowledge is elicited and capitalized among the participating EI actors in solving this problem. Decision maker presents the decision problem as initial demand and its contexts – internal and external environment using the interface shown in figure 7. This formulated initial demand is viewed by other participating actors as shown in figure 8. Subsequently, the watcher defines the stake of the DP using the annotation interface as illustrated in figure 9. Decision maker used annotation interface for the validation of the specified stake while the timestamp is also stored (figure 10). The watcher reacts to decision maker's annotation and update DP stake accordingly. This process is repeated until concession is reached. The knowledge of the subsequent tasks of the resolution process is capitalized with respective links on the EIKC system interface. Actors exploit the KM system by visualizing both validated and pre-validated or evolving knowledge of specific tasks of resolving DP according to EI process.

---

[1] http://www.informationnigeria.org/2010/03/neco-gce-results-released-98-of-candidates-fail-as-neco-releases-novdec.html (retrieved on 30/09/2010).

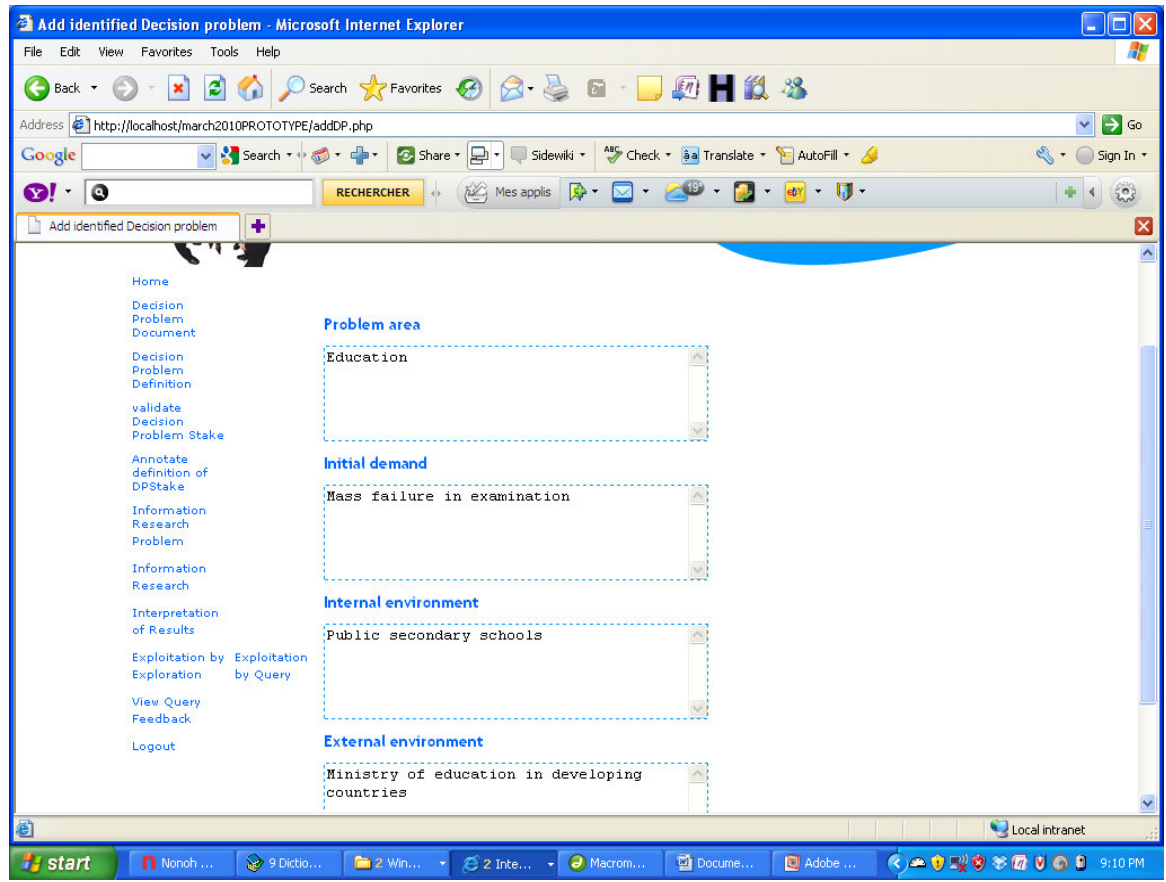

*Figure 7: Initial demand of DP*

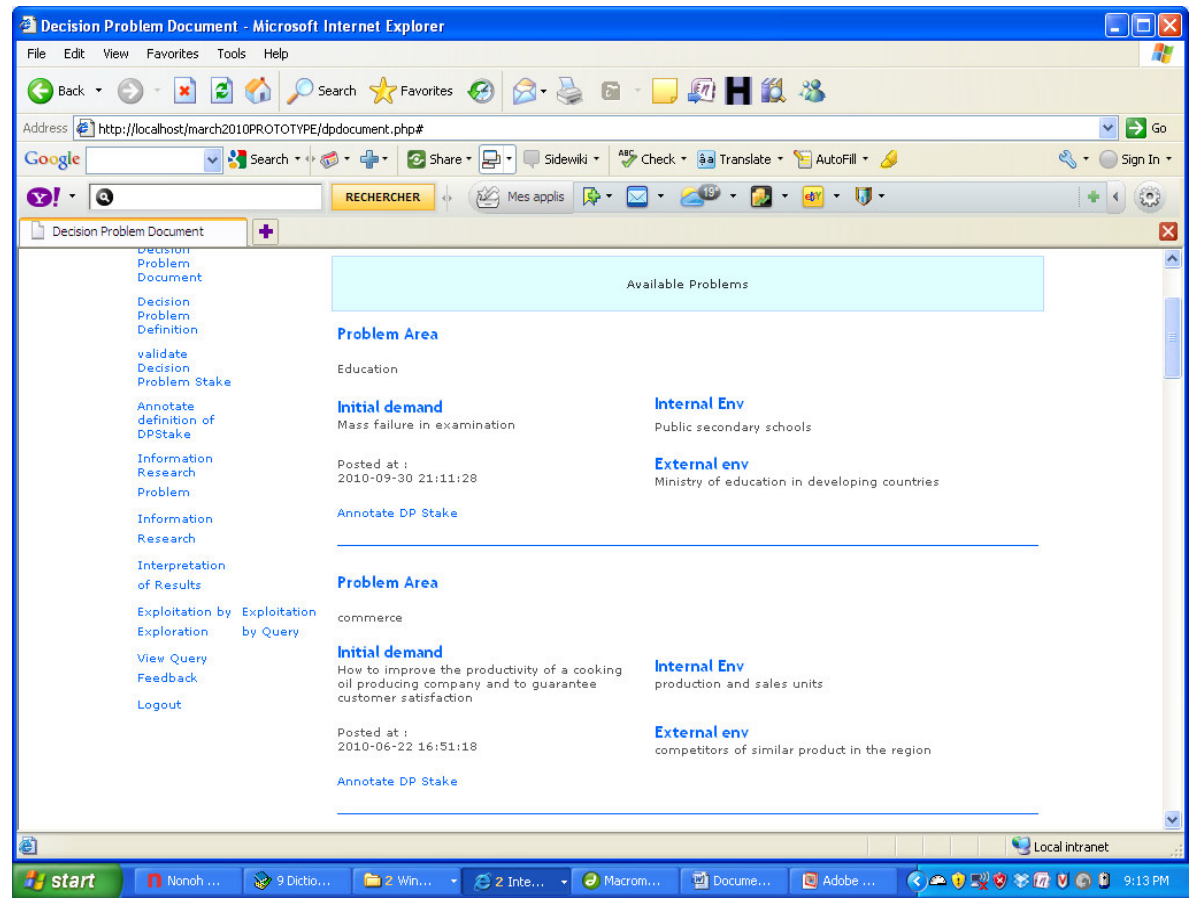

*Figure 8 : Visualization of expressed knowledge*

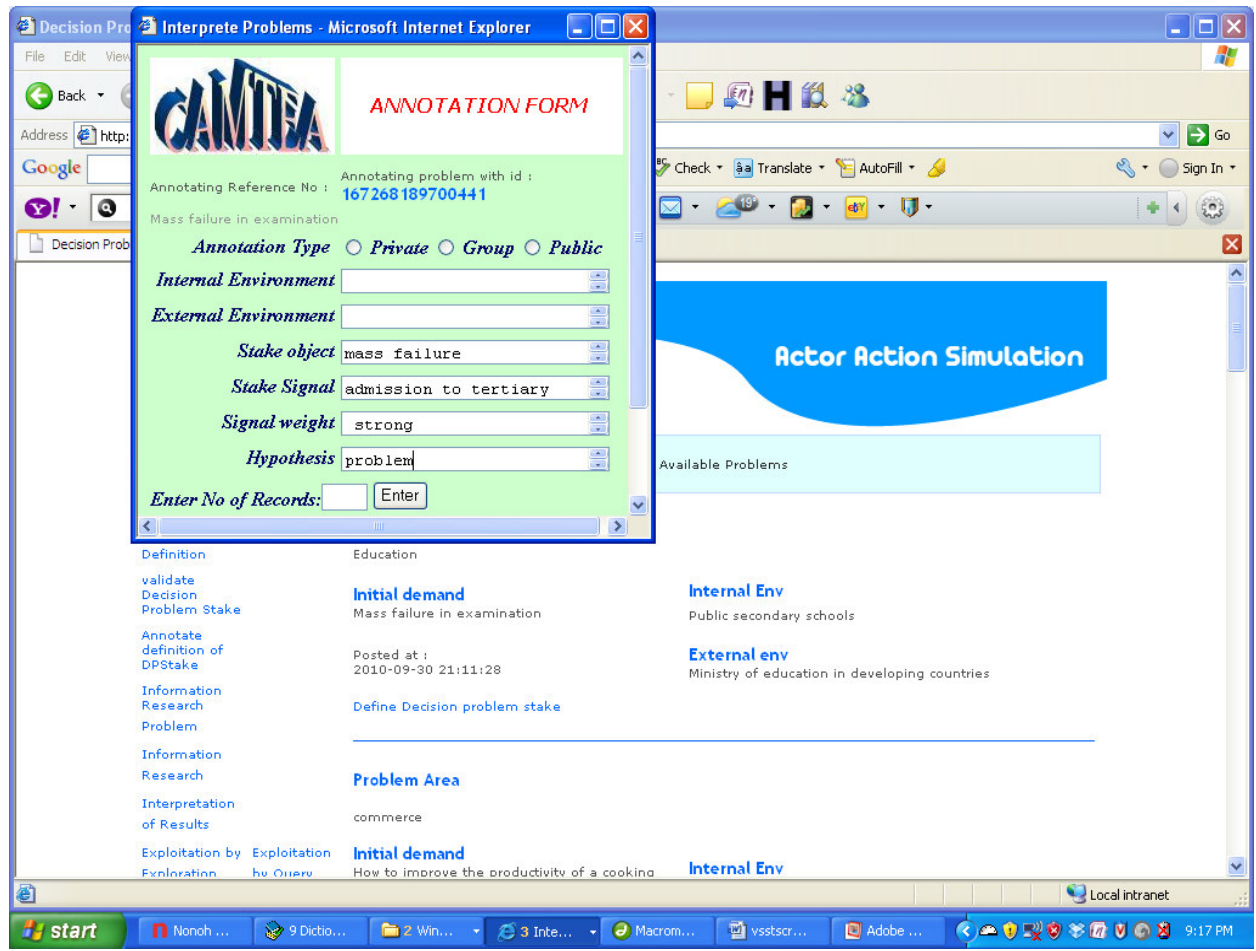

*Figure 9 : DP definition in terms of stake*

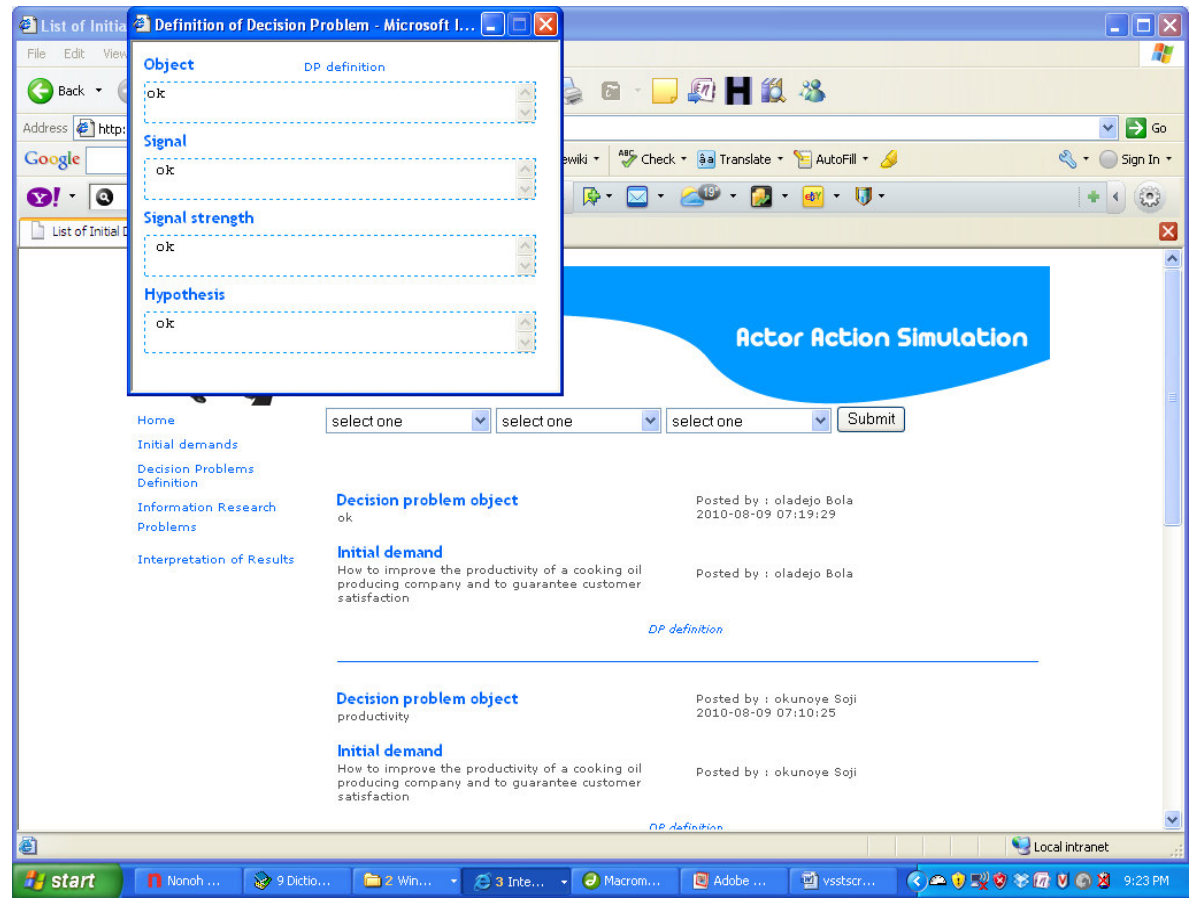

*Figure 10 : Validation through annotation interface*

The implication of the proposed capitalisation approach and EIKC system is that creation and exploitation of collaborative knowledge of actors in the course of handling DP resolution yield validated or reliable knowledge which is accessible to them based on their needs.

The second implemented system called MECOCIR handles the information process needed in resolving a decision problem. This system allows EI actors especially information watchers to engage in synchronous and asynchronous collaboration during information problem resolution phase. The actors can collaboratively suggest relevant information sources and as well search together for relevant information from such sources. The three level of group awareness that we proposed are fully implemented in this system. All the knowledge produced in the course of the interactions among the actors are as well captured and capitalized for future reuse. Figure 11 shows the main interface of MECOCIR.

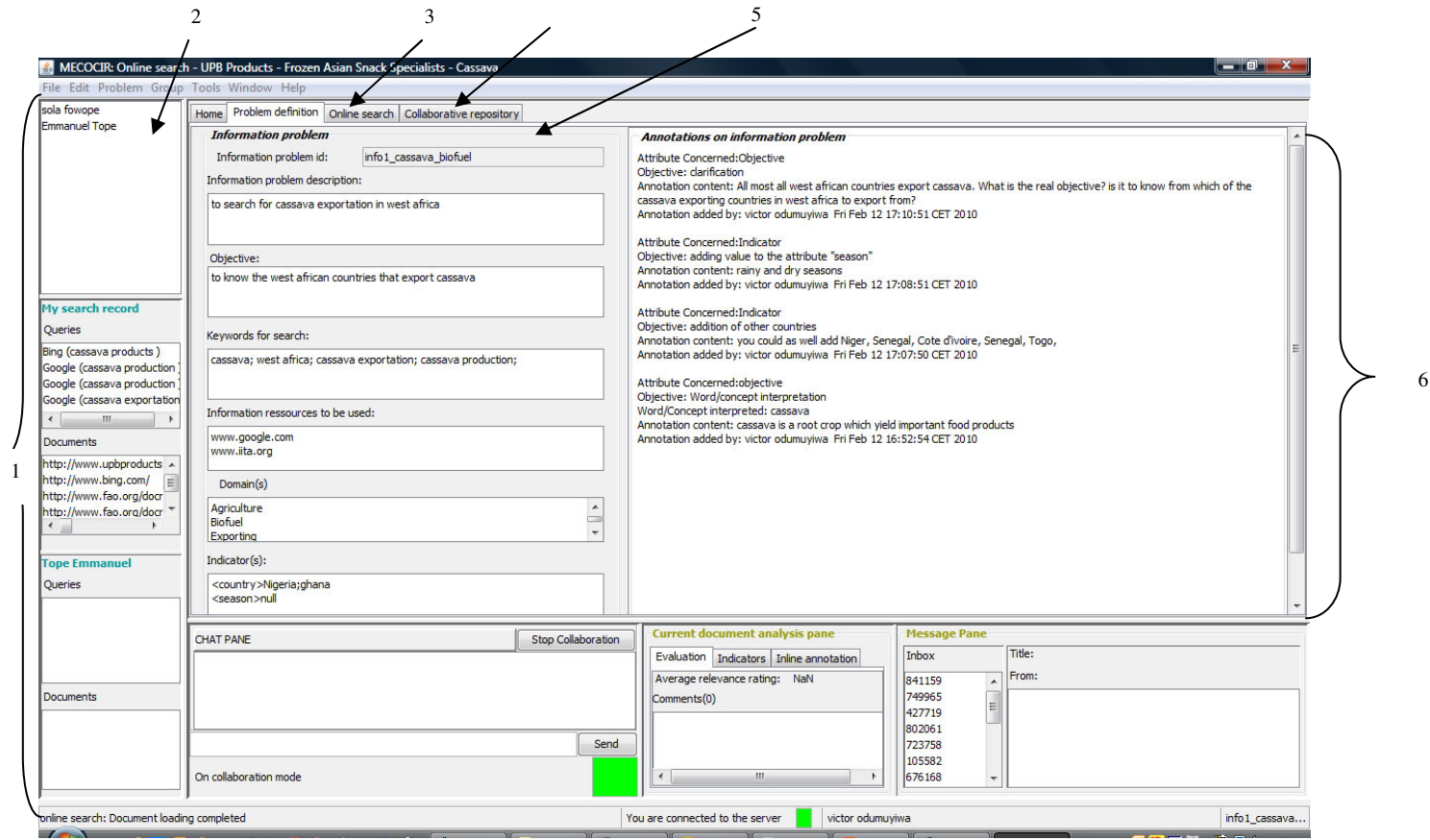

(1) Awareness interface, (2) Collaborating users list, (3) Browser interface tab for online search, (4) Collaborative repository interface, (5) Information problem definition interface, (6) Annotations on information problem

Figure 11: MECOCIR main interface

# 6. Conclusion

The importance of knowledge capitalisation in today's knowledge economy cannot be overemphasized. This paper discussed knowledge capitalisation approach in economic intelligence. We noted that EI actors collaborate in solving decision problem. During collaboration, there is need for each actor to express his knowledge in a form that can be shared and understood by other actors. We proposed an annotation model that allows for knowledge elicitation as attribute-value pair in EI. Expressed knowledge are capitalised using our dynamic capitalisation method. The need to manage interactions among collaborating actors led to the proposition of three level of group awareness. Our propositions were implemented in EIKC and MECOCIR system.

# 7. Reference

[1]. AFOLABI B. & THIERY O., Considering *users' behaviours in improving the responses of an information base*, in 1st International Conference on Multidisciplinary Information Sciences and Technologies, Merida, Spain, October, 2006.

[2]. BODAIN Y. & ROBERT J.M., *Developing a robust authoring annotation system for the Semantic Web*, Seventh IEEE International Conference on Advanced Learning Technologies (ICALT 2007) http://ieeexplore.ieee.org/iel5/4280926/4280927/04281043.pdf 20/10/2007

[3]. BRUSILOVSKY P., *Efficient technique for adaptive hypermedia, intelligent hypertext: Advances techniques for the World Wide Web*. Lecture notes in computer science, 1326, Berlin, Spring-Verlag, 12-30 1996

[4]. DAVID A. & THIERY O., *Application of "EQuA2te" Architecture in Economic Intelligence*. In Information and Communication Technologies applied to Economic Intelligence - ICTEI'2002, Ibadan, Nigeria, 2002.

[5]. DAVID A., BOUAKA N. & THIERY O., *Contribution to the understanding of explanatory factors for decision-maker problem within the framework of economic intelligence*, SCI' 2002, Orlando, Florida, USA, July 2002.

[6]. DOURISH P. & BELLOTTI V., *Awareness and coordination in shared workspaces*, In Proceedings of the ACM Conference on Computer Supported Cooperative Work (CSCW '92), p. 107-114, Toronto, Ontario, 1992, ACM Press

[7]. KISLIN P., *Modelisation du Problème Informationnel du Veilleur*, Thèse de doctorat, Université Nancy 2, France, November 2007

[8]. LONCHAMP J., *Le travail coopératif et ses technologies*, Lavoisier, Paris, 2003

[9]. MENENDEZ A. et al, *Economic intelligence. A guide for beginners and practitioners*, CETISME partnership, 2002. Retrieved March 21, 2008 from http://www.madrimasd.org/Queesmadrimasd/Socios_Europeos/descripcionproyectos/Documentos/CETISME-ETI-guide-english.pdf.

[10]. ODUMUYIWA V. & DAVID A., *Collaborative Information Retrieval among Economic Intelligence Actors*, in the fourth International Conference on Collaboration Technologies (CollabTech 2008), (pp. 21-26), Wakayama, Japan, August 2008.

[11]. ODUMUYIWA V. & DAVID A., *Collaborative knowledge creation and management in information retrieval*, in the 5th international KMO conference KMO' 2010, May 18 -19, 2010, (pp. 73-84), Veszprém, Hungary.

[12]. OKUNOYE O. B., DAVID A. & UWADIA C. *AMTEA: Tool for Creating and Exploiting Annotations in the Context of Economic Intelligence (Competitive Intelligence)*. IEEE IRI 2010 Conference, Las Vegas, USA; August 4 – 6, 2010.

[13]. OLADEJO B. F., ODUMUYIWA V. T. & DAVID A. A., *Dynamic capitalization and visualization strategy in collaborative knowledge management system for EI process*. International Conference of Knowledge Economy and Knowledge Management, 28-30 June, 2010, Paris

[14]. REVELLI C., *Intelligence stratégique sur Internet*, Paris, Dunod, 1998, pp. 18,19

[15]. ROBERT C., *L'annotation pour la recherche d'informations dans le contexte d'intelligence économique*, Thèse de doctorat, Université Nancy 2, février 2007